\documentclass[10pt,twocolumn,letterpaper]{article}

\usepackage{iccv}
\usepackage{times}
\usepackage{epsfig}
\usepackage{graphicx}
\usepackage{amsmath}
\usepackage{amssymb}
\usepackage{multirow}
\usepackage{subfigure}
\usepackage{makecell}
\usepackage{float}
\usepackage{comment}
\usepackage{blindtext}
\usepackage{graphicx}
\usepackage{caption}
\usepackage{booktabs}

\usepackage[pagebackref=true,breaklinks=true,letterpaper=true,colorlinks,bookmarks=false]{hyperref}

\iccvfinalcopy 

\def\iccvPaperID{8847} 

\ificcvfinal\fi

\usepackage[T1]{fontenc}
\usepackage[utf8]{inputenc}
\usepackage{authblk}

\title{DeepMultiCap: Performance Capture of Multiple Characters \\ Using Sparse Multiview Cameras\vspace{-25pt}}

\author{Yang Zheng$^*$, Ruizhi Shao$^*$, Yuxiang Zhang, Tao Yu, Zerong Zheng, Qionghai Dai, Yebin Liu
\\Department of Automation and BNRist, Tsinghua University}

\begin{document}

\twocolumn[{%
\renewcommand\twocolumn[1][]{#1}%
\maketitle
\begin{center}
    \centering
    \vspace{-18pt}
    \includegraphics[width=\textwidth]{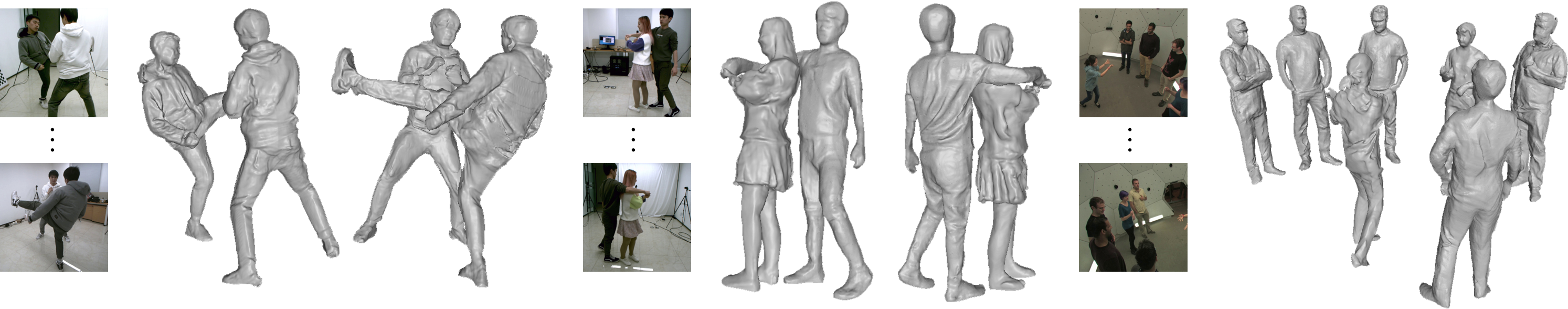}
    \vspace{-17pt}
    \captionof{figure}{
    Given only sparse multi-view RGB videos (6 views for the left and middle, 8 views for the right), our method is able to reconstruct various kinds of 3D shapes with temporal-varying surface details even under challenging occlusions for multi-person interactive scenarios. 
    }
    \label{teaser}
\end{center}%
}]
\begin{abstract}
We propose DeepMultiCap, a novel method for multi-person performance capture using sparse multi-view cameras. Our method can capture time varying surface details without the need of using pre-scanned template models. To tackle with the serious occlusion challenge for close interacting scenes, we combine a recently proposed pixel-aligned implicit function with parametric model for robust reconstruction of the invisible surface areas. 
An effective attention-aware module is designed to obtain the fine-grained geometry details from multi-view images, where high-fidelity results can be generated.
In addition to the spatial attention method, for video inputs, we further propose a novel temporal fusion method to alleviate the noise and temporal inconsistencies for moving character reconstruction. 
For quantitative evaluation, we contribute a high quality multi-person dataset, MultiHuman, which consists of 150 static scenes with different levels of occlusions and ground truth 3D human models.
Experimental results demonstrate the state-of-the-art performance of our method and the well generalization to real multiview video data, which outperforms the prior works by a large margin.
\end{abstract}

\let\thefootnote\relax\footnotetext{* Equal contribution}
\footnotetext{Code and dataset available: \href{http://liuyebin.com/dmc/dmc.html}{http://liuyebin.com/dmc/dmc.html}}
\section{Introduction}
Recent years have witnessed great progress in vision-based human performance capture, which is promising to enable various applications (e.g., tele-presence, sportscast, gaming and mixed reality) with enhanced interactive and immersive experiences. 
To achieve surprisingly detailed geometry and texture reconstruction, dense camera rigs even equipped with sophisticated lighting systems are introduced ~\cite{vlasic2009dynamic, collet2015high, joo2018total,joo2019panoptic,guo2019relightables,bagautdinov2021driving}. However, the extremely expensive and professional setups limited their popularity. 
Although other light-weight multi-view human performance capture systems have achieved impressive results even in real-time, they still relies on pre-scanned templates~\cite{liu2011markerless,liu2013markerless}, custom-designed RGBD~\cite{dou2016fusion4d,dou2017motion2fusion} or commercial RGBD~\cite{yu2021function4d, yu2018doublefusion} sensors, or limited to single-person reconstruction~\cite{Starck07,gall2009motion,huang2018deep,Minimal18}. 

Benefiting from the fast improvement of deep implicit functions for 3D representations, recent methods~\cite{saito2019pifu,saito2020pifuhd,Monoport2020} are able to recover the 3D body shape only from a single RGB image. 
Compared with the voxel-based~\cite{varol2018bodynet, zheng2019deephuman} or mesh-based~\cite{natsume2019siclope, alldieck2019tex2shape} representations, 
an implicit function guides the deep learning models to notice geometric details in a more efficient way. 
Specifically, PIFu~\cite{saito2019pifu,Monoport2020} achieves plausible single human reconstruction using only RGB images, and PIFuHD~\cite{saito2020pifuhd} further utilizes normal maps and high resolution images to generate more detailed results. 

Despite the prominent performance in digitizing 3D human body, both PIFu~\cite{saito2019pifu} and PIFuHD~\cite{saito2020pifuhd} suffer from several drawbacks when extending the frameworks to multi-person scenarios and multi-view setups. 
Firstly, the average-pooling-based multi-view feature fusion strategy in PIFu will lead to over-smoothed outputs when high frequency details (e.g., normal maps) are included in multi-view features.
More importantly, in the two approaches, reconstruction results are only promised with ideal input images without severe occlusions in multi-person performance capture scenarios. 
The reconstruction performance of~\cite{saito2019pifu,saito2020pifuhd} will be significantly deteriorated due to the lack of observations caused by severe occlusions. 

To address the aforementioned problems, we propose a novel framework to perform multi-person reconstruction from multi-view images. 
First of all, inspired by~\cite{vaswani2017attention}, we design an spatial attention-aware module to adaptively aggregate information from multi-view inputs.
The module is effective to capture and merge the geometric details from different view points, and finally contributing to the significant improvement of results under multi-view setups. 
Moreover, for multi-person reconstruction, we further combine the attention module with parametric models, i.e., SMPL to enhance the robustness while maintaining the fine-grained details. 
The SMPL model serves as a 3D geometry proxy which compensates for the missing information where occlusions take place. 
With the semantic information provided by SMPL, the network is capable of reconstructing complete human bodies even under close interactive scenarios.
Finally, when dealing with moving characters from video, we propose a temporal fusion method by weighting the signed distance field (SDF) across the time domain, which further enhance the temporal consistency of the reconstructed dynamic 3D sequences.

Another urgent problem is that the lack of high-quality scans of multi-person interactive scenarios in the community makes it difficult for accurately evaluating multi-person performance capture systems like ours. 
To fill this gap and better evaluate the performance of our system, we contribute a novel dataset, MultiHuman, which consists of 150 high-quality scans with each containing from 1 to 3 multi-person interactive actions (including both natural and close interactions). 
The dataset is further divided into several categories according to the level of occlusions and number of persons in the scene, where a detailed evaluation can be conducted.
Experimental results demonstrate the state-of-the-art performance and well generalization capacity of our approach. 
In general, the main contribution in this work can be summarized as follows:
\begin{itemize}
\item We propose a novel framework for high-fidelity multi-view reconstruction for multi-person interactive scenarios. By leveraging the human shape and pose prior for resolving the ambiguities introduced by severe occlusions, we achieve the state-of-the-art performance even with partial observations in each view.
\item We design an efficient spatial attention-aware module to obtain fine-grained details for multi-view setups, and introduce a novel temporal fusion method to reduce the reconstruction inconsistencies for moving characters from video inputs.
\item We contribute an extremely high quality 3D model dataset containing of 150 multi-person interacting scenes. The dataset can be used for training and evaluation of related topics in future research. 
\end{itemize}

\section{Related Work}
\noindent\textbf{Single-view performance capture}
Many methods have been proposed to reconstruct detailed geometry from single-view inputs. 
Typical techniques include silhouette estimation~\cite{natsume2019siclope}, depth estimation~\cite{gabeur2019moulding, smith2019facsimile} and template-based deformation~\cite{alldieck2019tex2shape, zhu2019detailed, habermann2020deepcap}. 
Moreover, SMPL~\cite{loper2015smpl} regression or optimization can be incorporated to generate more reliable and robust outputs as shown in~\cite{zheng2019deephuman, zheng2021pamir, bhatnagar2020combining}. 
Real-time methods can be implemented with the aid of a single depth sensor~\cite{yu2017bodyfusion, yu2018doublefusion} or by innovating computation and rendering algorithms~\cite{Monoport2020}. 
Regarding to the 3D representations used in these methods, we can split them into two categories: explicit~\cite{varol2018bodynet, natsume2019siclope, zheng2019deephuman, he2021challencap} and implicit~\cite{saito2019pifu, saito2020pifuhd, huang2018deep, huang2020arch, chibane2020implicit, deng2020nasa, bhatnagar2020loopreg, Wang_2021_CVPR, shao2021doublefield} reconstruction methods. 
Compared with traditional explicit 3D representations, implicit representations show certain advantages in domain-specific shape learning and detail preservation.
For example, PIFu~\cite{saito2019pifu} define the surface as a level set of function $f$. Similarly,~\cite{huang2018deep} defines a
probability field of surface points, and ARCH~\cite{huang2020arch}
predicts a 3D occupancy map.
However, all of the methods above are mainly focusing on single-person reconstruction, and it remains difficult for them to achieve accurate reconstruction under multi-person scenarios. 

\begin{figure*}[t]
   \centering
   \vspace{-20pt}
   \includegraphics[width=\linewidth]{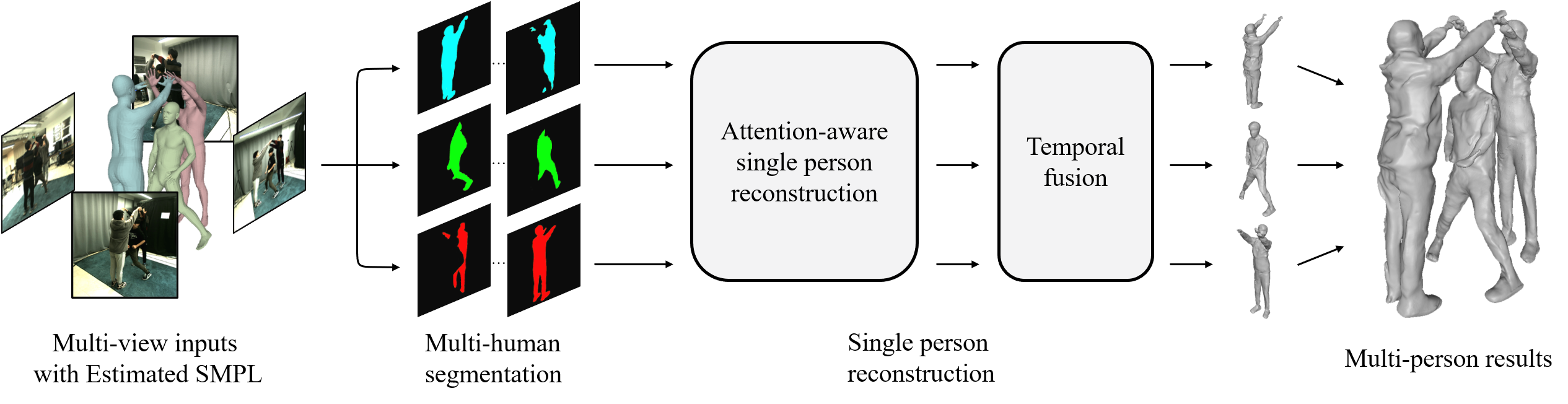}
   \vspace{-20pt}
   \caption{Pipeline of our system. With estimated SMPL models and segmented multi-view, we design a spatial attention-aware network and temporal fusion method to 
   reconstruct each character separately. 
   }
   \vspace{-15pt}
   \label{fig:pipeline}
\end{figure*}

\noindent\textbf{Multi-view performance capture}
Motion capture has been developed to make accurate motion predictions in multi-person interaction scenes~\cite{belagiannis20143d, liu2011markerless, liu2013markerless, joo2019panoptic, rogez2019lcr, rogez2017lcr, lightcap2021, pymaf2021, kwon2020recursive, ohashi2020synergetic, tu2020voxelpose, lin2021multi, Zanfir_2018_CVPR}. Some of them even achieve real-time performance~\cite{bridgeman2019multi, dong2019fast, zhang20204d, mehta2020xnect}. 
However, these works only capture skeleton motions instead of the detailed geometries.
Regarding to multi-view geometry reconstruction, previous studies use template-based deforming methods~\cite{de2008performance, vlasic2008articulated, gall2009motion}, skeleton tracks~\cite{vlasic2008articulated, gall2009motion} or fusion-based techniques~\cite{dou2013scanning}. 
Aside from the long computation time, these methods often show deficiency in mapping textures, handling topology changes or dealing with drastic frame-to-frame motion. 
Moreover, the aforementioned methods also show limited adaptability for multi-person capture as they cannot effectively deal with occlusions.
Robust quality reconstruction methods often come at prohibitive dependencies and constraints. 
Some methods depend on dense viewpoints~\cite{collet2015high, joo2018total} and even controlled lighting~\cite{vlasic2009dynamic,guo2019relightables} to reconstruct detailed geometry. 
Another branch of multi-view RGBD systems~\cite{dou2016fusion4d, dou2017motion2fusion, yu2021function4d, zhi2020texmesh, pang2021few}, achieve impressive real-time performance capture results even for multi-person scenarios benefiting from the strong depth observations. 
Note that Huang et,al.~\cite{huang2018deep} also presents a volumetric capture approach to accomplish quality results using very sparse-view RGB inputs, but they only focus on single-person reconstruction without considering how to resolve the challenges introduced by multi-person occlusions. 

\noindent\textbf{Attention-based network}
Apart from the huge success of attention mechanism in natural language processing~\cite{vaswani2017attention}, attention-based network has achieved prominent performance in visual tasks, including image classification~\cite{wang2017residual}, 
image segmentation~\cite{zhang2018context, yu2018learning, li2019attention}, super-resolution~\cite{dai2019second}, 
multi-view stereo~\cite{luo2020attention} and hand pose estimation~\cite{huang2020hot}. 
In these works, attention mechanism is applied to capture the correlation of embedding features or context relationship of hierarchical structure. In particular, Luo \textit{et al.}~\cite{luo2020attention} propose an attention-aware network AttMVS to merge contextual information from multi-view scenes. An attention-guided regularization module is used for more robust prediction. In~\cite{huang2020hot}, Lin \textit{et al.} design a non-autoregressive transformer to learn the structural correlations among hand joints. Recent findings~\cite{naseer2021intriguing} have demonstrated that the self-attention mechanism~\cite{vaswani2017attention} is highly robust to severe occlusions in vision tasks.
\section{Overview}
An overview of our approach is illustrated in Fig~\ref{fig:pipeline}. With input as the segmented multi-view single person images and the corresponding SMPL, the system outputs the reconstructed 3D persons. 
The results are combined together directly with no need for modifying the relative position, since the multi-view setting ensures the 3D spatial relationship between different individuals. 

To obtain the inputs, we firstly fit SMPL-X~\cite{pavlakos2019expressive} models through 3D keypoints estimated from multi-view video by a light-weight total capture method~\cite{lightcap2021}.
For multi-person segmentation, we refer to a self-correction method~\cite{li2020self} and use SMPL projection maps to track different characters in multi-view scenes.
Finally, the 3D human can be generated through the spatial attention-aware network based on the pixel-aligned implicit function, and further polished by the temporal fusion method when the time information is available in the video inputs, which will be described detailedly in Section~\ref{sec 4}. 

\begin{figure*}[t]
   \centering
   \vspace{-10pt}
   \includegraphics[width=\linewidth]{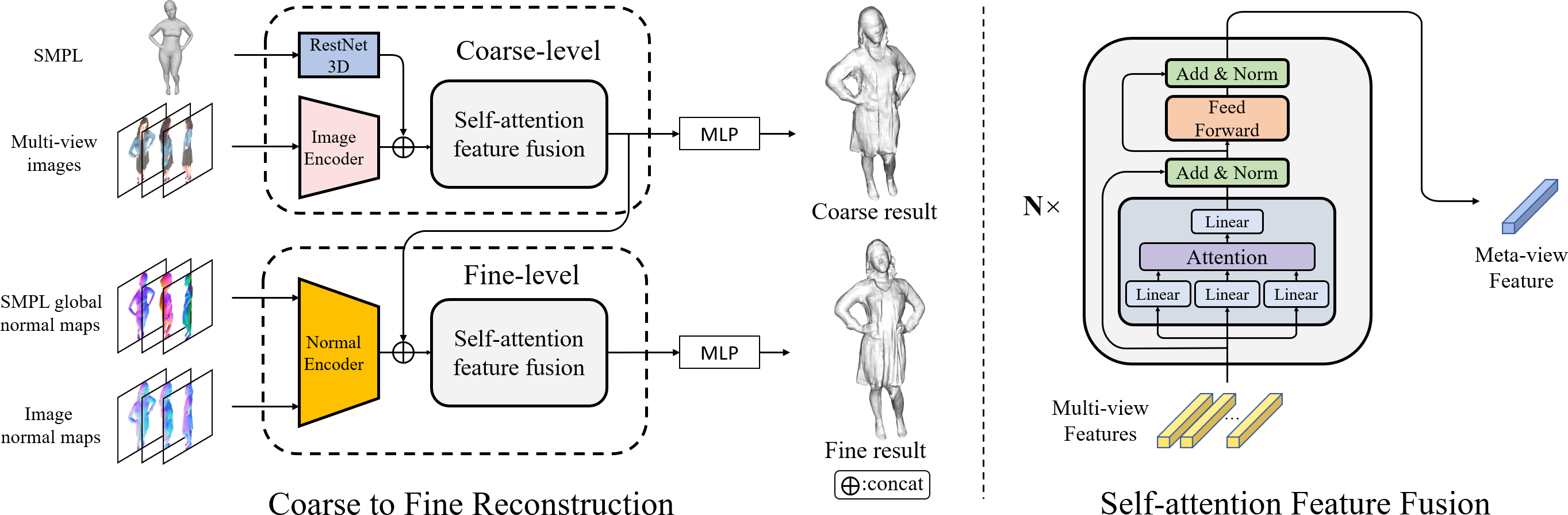}
   \vspace{-10pt}
   \caption{Architecture of our attention-aware network. 
   We leverage a two-level coarse to fine framework (left) with a multi-view feature fusion module based on self-attention (right). 
   Human body prior SMPL is used in the coarse level to ensure the robustness of reconstruction, 
   and a specially designed SMPL global normal map helps the fine level network better capture the details. 
   To merge multi-view features efficiently, we leverage the self-attention mechanism to extract meta information from different observations, which significantly improve the reconstruction quality.
   }
   \vspace{-10pt}
   \label{fig:overview}
\end{figure*}
 
\subsection{Preliminary}
 \label{sec:pre}
 Our method is based on the implicit function. An implicit function represents the surface of a 3D model as a level set of an occupancy field function F, e.g. 
 F(X) = 0.5. Specifically, 
 PIFu~\cite{saito2019pifu} combines 3D points with conditional variables to formulate a pixel-aligned implicit function:
 \begin{equation}
    \label{pifu}
    F(\Phi(\boldsymbol{x}, \boldsymbol{I}), z(\boldsymbol{X}))=s:s\in[0, 1]
 \end{equation}
 where for an image $\boldsymbol{I}$ and a given 3D point $\boldsymbol{X}$,
  $\boldsymbol{x}=\Pi(\boldsymbol{X})$ is the 2D projection coordinate on the image plane, 
  $z(\boldsymbol{X})$ is the depth value in the camera space, and $\Phi(\boldsymbol{x}, \boldsymbol{I})$ is the image embedding feature at location $\boldsymbol{x}$.
 In PIFu, a multi-layer perceptron (MLP) is trained to fit F. 
 
In order to improve the quality of the reconstruction results, PIFuHD~\cite{saito2020pifuhd} maintains the origin 
 PIFu framework as a coarse level prediction while adding high resolution images to a fine level network:
 \begin{equation}
    \label{pifuhd}
    F^{H}(\Phi(\boldsymbol{x}, \boldsymbol{I}_H, \boldsymbol{N}_F, \boldsymbol{N}_B), \Omega(\boldsymbol{X}))=s:s\in[0, 1]
 \end{equation}
 where $\boldsymbol{I}_H, \boldsymbol{N}_F, \boldsymbol{N}_B$ are the high resolution image, the predicted frontal and back normal map, and $\Omega(\boldsymbol{X})$ is the 3D embeddings extracted from the intermediate features in the coarse level. 
 More detailed human models can be reconstructed with additional information brought by the increasing resolution and high frequent details in the normal maps.
 
For multi-view images, a naive strategy is proposed in PIFu~\cite{saito2019pifu} to extract multi-view features, i.e., performing mean pooling on the embeddings from the intermediate layer of the MLP. 
However, this simple method may lead to loss of details and even collapse in real world cases, especially when the multi-view features are not consistent due to the various depth in different views and occlusions. 

\section{Single Person Reconstruction}
\label{sec 4}
Reconstruction of a single person from multi-view is a challenging problem. The main concern is to extract the meta information of the observations from different views.
For this end, we propose an novel feature fusion module based on self-attention mechanism, which is effective to help the network aware of geometry details shown in the multi-view scenes. 
To tackle with the inconsistencies and loss of information brought by occlusions, we combine the attention module with parametric models to enhance the robustness of reconstruction while preserving the fine-grained details.  
The architecture of our network is illustrated in Figure~\ref{fig:overview}. 
Following PIFuHD~\cite{saito2020pifuhd}, our method builds on a coarse-to-fine framework. 
The coarse level conditioned with images and SMPL models ensures a confident result, 
and the fine level refines the reconstruction by utilizing high resolution image feature maps ($512\times512$). 
The results can be further polished by a temporal fusion method when time information is available for video inputs.
With the proposed spatial attention and temporal fusion framework, the reconstruction remains robust and in high quality . 

\subsection{Attention-aware Multi-view Feature Fusion}
\label{sec3:attention}
In PIFu~\cite{saito2019pifu}, the simple strategy for multi-view reconstruction is averaging the multi-view feature embeddings from the intermediate layer of MLP.
We argue that the method is not efficient enough to merge the geometry details from multi-view scenes, which could lead to losing information.
As shown in Figure~\ref{fig:view consistency}, when the strategy is applied to PIFuHD~\cite{saito2020pifuhd}, we obtain a smoother output. 
The geometry features may not remain consistent since the visible regions changes from view to view. The mean pooling method cannot handle these cases effectively.

To capture correlations between different views, inspired by~\cite{vaswani2017attention}, we propose a multi-view feature fusion method based on self-attention mechanism. 
The detailed architecture of the module is illustrated in Figure~\ref{fig:overview}. 
In practice, given $n$ observations, the multi-view features are stacked together as $\boldsymbol{\phi}_m$, which is then embedded with three different linear layers and self-attention mechanism is applied:
\begin{equation}
    attention(\boldsymbol{\phi}_q, \boldsymbol{\phi}_s, \boldsymbol{\phi}_t)=softmax(\frac{\boldsymbol{\phi}_q^T\boldsymbol{\phi}_s}{\sqrt{d_k}})\boldsymbol{\phi}_t
\end{equation}
where $\boldsymbol{\phi}_q=\boldsymbol{\phi}_m\boldsymbol{W}_q, \boldsymbol{\phi}_s=\boldsymbol{\phi}_m\boldsymbol{W}_s, \boldsymbol{\phi}_t=\boldsymbol{\phi}_m\boldsymbol{W}_t$ denote the 
query, source and target feature embedded by learnable weights $\boldsymbol{W}$, and $\boldsymbol{\phi}_m \in \mathbb{R}^{n\times d_k}$ with $d_k$ as the embedding size. 
The dot-product result is divided by $\sqrt{d_k}$ to prevent the gradient vanishing problem.
For linear weights $\boldsymbol{W}$, we use 
a multi-head attention strategy~\cite{vaswani2017attention}, i.e, $\boldsymbol{W}_q, \boldsymbol{W}_s, \boldsymbol{W}_t \in \mathbb{R}^{n_{head}\times d_k}$ which encode the multi-view features into $n_{head}$ 
different embedding subspaces, allowing the model to better notice different geometry patterns jointly. 
As the result, the weights of $n$ observations are obtained through 
\emph{softmax} function by calculating the similarity between views in the query feature $\boldsymbol{\phi}_q$ and the source feature $\boldsymbol{\phi}_s$. 
The salient covisible details tend to have large weights and will be maintained, 
while the invisible regions which lead to small weights have little influence on the outputs.

Finally, we stack the linear and attention layers to form a self-attention encoder as proposed in~\cite{vaswani2017attention}. 
The meta-view prediction is then generated as:
\begin{equation}
  \label{feature fusion}
  F^T(\boldsymbol{X}) = g^T(T(\boldsymbol{\phi}_m)) 
\end{equation}
where $T(\boldsymbol{\phi}_m)$ is the feature output of the self-attention encoder, and the implicit function $g^T$ predicts the occupancy field. 
The output meta-view feature is expected to contain the global spatial information. 
As demonstrated in Figure~\ref{fig:view consistency}, when combining the attention module with PIFuHD~\cite{saito2020pifuhd}, we are able to capture and preserve details with increasing observations.

\subsection{Embedded with Parametric Body Model}
\label{sec:smpl}
Although the attention-aware feature fusion module is effective to mine details from multi-view, without auxiliary 3D information,
the network struggles to make a reasonable prediction when information is lost due to occlusions. 
To address the limitation, we combine the strength of attention mechanism and parametric models. 

A parametric body model, e.g, SMPL, contains the pose and shape information of human bodies. 
The semantic feature of SMPL is extracted by 3D convolution network for geometry inference. 
To improve the efficiency of attention module, inspired by the position encoding introduced in~\cite{vaswani2017attention}, we further design an informative view representation by rendering SMPL global normal maps. 
The global maps offer guidance for the network to identify the particular visible body parts in multi-view observations, 
and the corresponding geometry features can be easily extracted. 
Specially, to render the global normal maps, SMPL is transformed to the canonical model space, 
where RGB color is obtained from the normal vector and standard rendering procedure can be applied. 
In multi-person scenes, though images of single person can be fragmentary due to occlusions, 
the extra information provided by SMPL compensates for the missing part and remain consistent under different views, which significantly improves the 
quality and robustness of reconstruction results.

With SMPL, we rewrite the two-level pixel-aligned function Eqn.~\ref{pifu} and Eqn.~\ref{pifuhd}. The coarse level has the formulation:
\begin{equation}
   \label{coarse_level}
   F^L(\boldsymbol{X}) = g^L(\Phi^L(\boldsymbol{x}, \boldsymbol{I}), \Psi(\boldsymbol{X}, \boldsymbol{V}_M))
\end{equation}
and the fine level:
\begin{equation}
   \label{fine_level}
   F^H(\boldsymbol{X}) = g^H(\Phi^H(\boldsymbol{x}, \boldsymbol{I}, \boldsymbol{N}_F, \boldsymbol{N}_{V_M}), \Omega^L(\boldsymbol{X}))
\end{equation}
where $\boldsymbol{V}_M$ denotes the volumetric representation of SMPL, $\Psi(\boldsymbol{X}, \boldsymbol{V}_M)$ is the SMPL semantic features, 
$\boldsymbol{N}_F, \boldsymbol{N}_{V_M}$ refer to the 
predicted frontal normal map and the rendered SMPL global normal map, and $\Omega^L(\boldsymbol{X})$ is the 3D embeddings from the coarse level. 
Note that here we perform the reconstruction under the SMPL model space where SMPL is normalized to a unit cube, allowing for the same setups during training and inference.
The reconstruction results are then transformed to the world space and aligned together. 

\subsection{Temporal Fusion}
For moving characters in video inputs, inconsistencies could raise between continuous frames due to the change of visible parts. 
To address the limitation, we propose a simple temporal fusion method. 
Suppose $p_{i, t}$ is a vertex of the reconstructed mesh at time $t$, we first calculate the blending weight by:
\begin{figure}[ht!]
   \centering  
   \includegraphics[width=\linewidth]{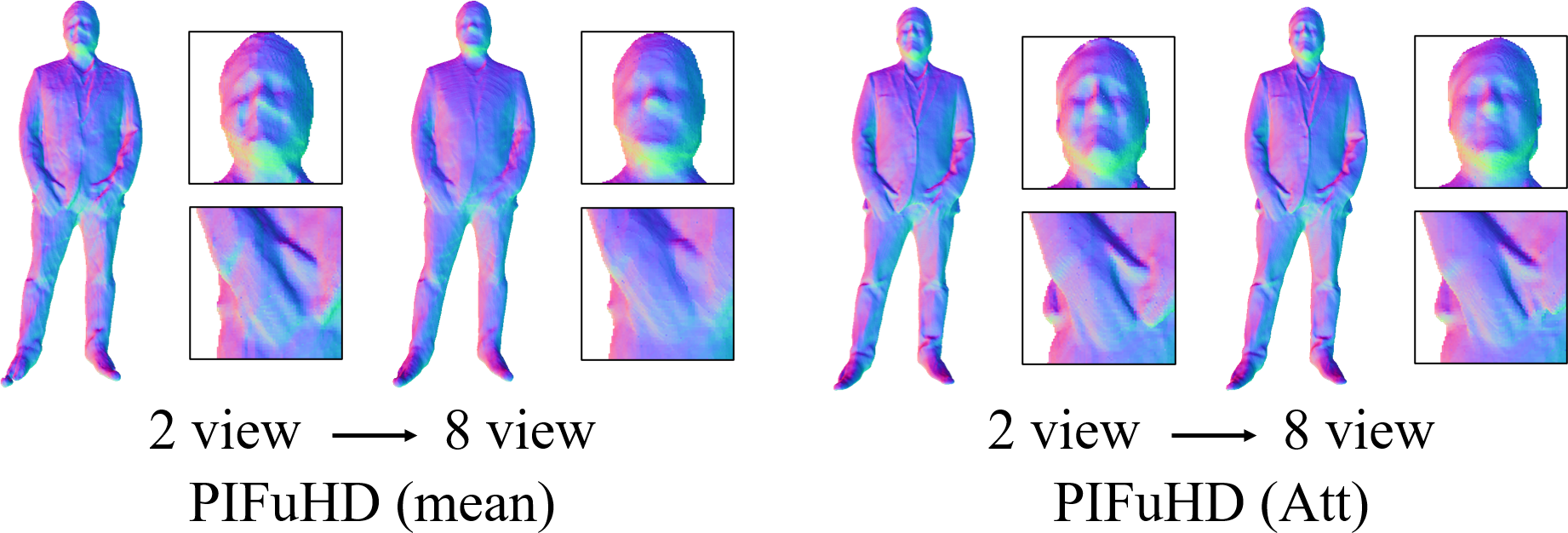}  
   \vspace{-15pt}
   \caption{When extending PIFuHD~\cite{saito2020pifuhd} to multi-view, mean pooling method (left) leads to smoother outputs while the attention module (right) helps to preserve the details.}
   \vspace{-10pt}
   \label{fig:view consistency}
\end{figure}
\begin{equation}
   \mathbf{W}_{t, i} = \sum_{j \in \mathcal{N}_{t, i}} \frac{w_{t, j\rightarrow i}}{w_{t, i}}\mathcal{W}_j
\end{equation}
where $\mathcal{N}_{t, i}$ is the nearest SMPL vertex set of $p_{t, i}$, $\mathcal{W}_j$ is the blending weight of SMPL vertex $v_{j, t}$, and
\begin{equation}
    w_{t, j\rightarrow i} = \exp(-\frac{\|p_{i, t}-v_{j, t}\|}{\sigma ^2}), w_{t, i} = \sum_{j \in \mathcal{N}_{t, i}}w_{t, j\rightarrow i}
\end{equation}
is the weight of vertex $v_{j, t}$. Given the estimated SMPL models at time $t$ and $t'$, the reconstructed vertices $V_t$ can then be warped to time $t'$ through the standard blend skinning:
\begin{equation}
    V_{t'\leftarrow t} = W(W^{-1}(V_{t}, J(\beta_t), \theta_t, \mathbf{W}_t), J(\beta_{t'}), \theta_{t'}, \mathbf{W}_t)
\end{equation}
where $W$ refers to the skinning procedure, and $J, \beta, \theta$ are the SMPL parameters. 
With the warped mesh, we calculate the signed distance field (SDF) and perform mean pooling to generate continuous reconstructions:
\begin{equation}
   S_{fusion}(t) = \frac{1}{h} \sum_{t' \in \mathcal{F}_t} S(t \leftarrow t')
\end{equation}
where $S$ denotes the SDF, and $\mathcal{F}_t$ is a sliding time window with size of $h$. In our approach, $h$ is set to 3 for consistent results while maintaining details.

\section{Extend to Multi-person Reconstruction}
\label{sec 5}
Multi-person reconstruction is implemented by reconstructing each individual separately.
The key challenge is to train the network to maintain robust against occlusions in interactive scenes.
For this end, we add handcrafted occlusions during training.
We firstly collect 1700 single human models from Twindom\footnote[1]{https://web.twindom.com/} and THuman2.0~\cite{yu2021function4d} to construct a large scale dataset.
To simulate multi-person cases, we render images via taichi~\cite{hu2019taichi} and randomly project other persons to the masks, 
where various situations can be generated from non-occlusion to heavy occlusions.
The high quality 3d models in the dataset enable us to render photorealistic pictures, and thus the network can be trained with well generalization ability to real world data.

\section{Dataset and Experiment}
In this section we explain our experimental settings and results. 
We highly recommend readers to refer to the supplementary document and video for more implementation details and a better visualization of our dataset and results.  
\subsection{MultiHuman Dataset}
Current human dataset only provides single person scans (Twindom and THuman2.0~\cite{yu2021function4d}), or 3D skeletons and deformable parametric models~\cite{joo2018total, Joo_2015_ICCV}.
The lack of high-quality scans of multi-person scenes limit the developement of the community.
For this end, we propose MultiHuman dataset, which is collected using a dense camera-rig equipped with 128 DLSRs and a commercial photogrammetry software. This system has also been used in constructing THuman2.0~\cite{yu2021function4d}.
Our dataset contains 150 multi-person static scenes. In total there are 278 characters with mostly university students wearing casual clothes. Each scene contains 1 to 3 persons, where each model consists of about 300,000 triangles with photorealistic texture.

To evaluate our method, we divide the dataset into different categories by the level of occlusions and number of persons, i.e., 30 single human scenes, 18 occluded single human scenes (by different objects), 46 natural interactive two person scenes, 30 closely interactive two person scenes, 
and 26 scenes with three persons. 

\subsection{Evaluation}

\begin{figure*}[ht]
   \centering
   \vspace{-15pt}
   \includegraphics[width=\linewidth]{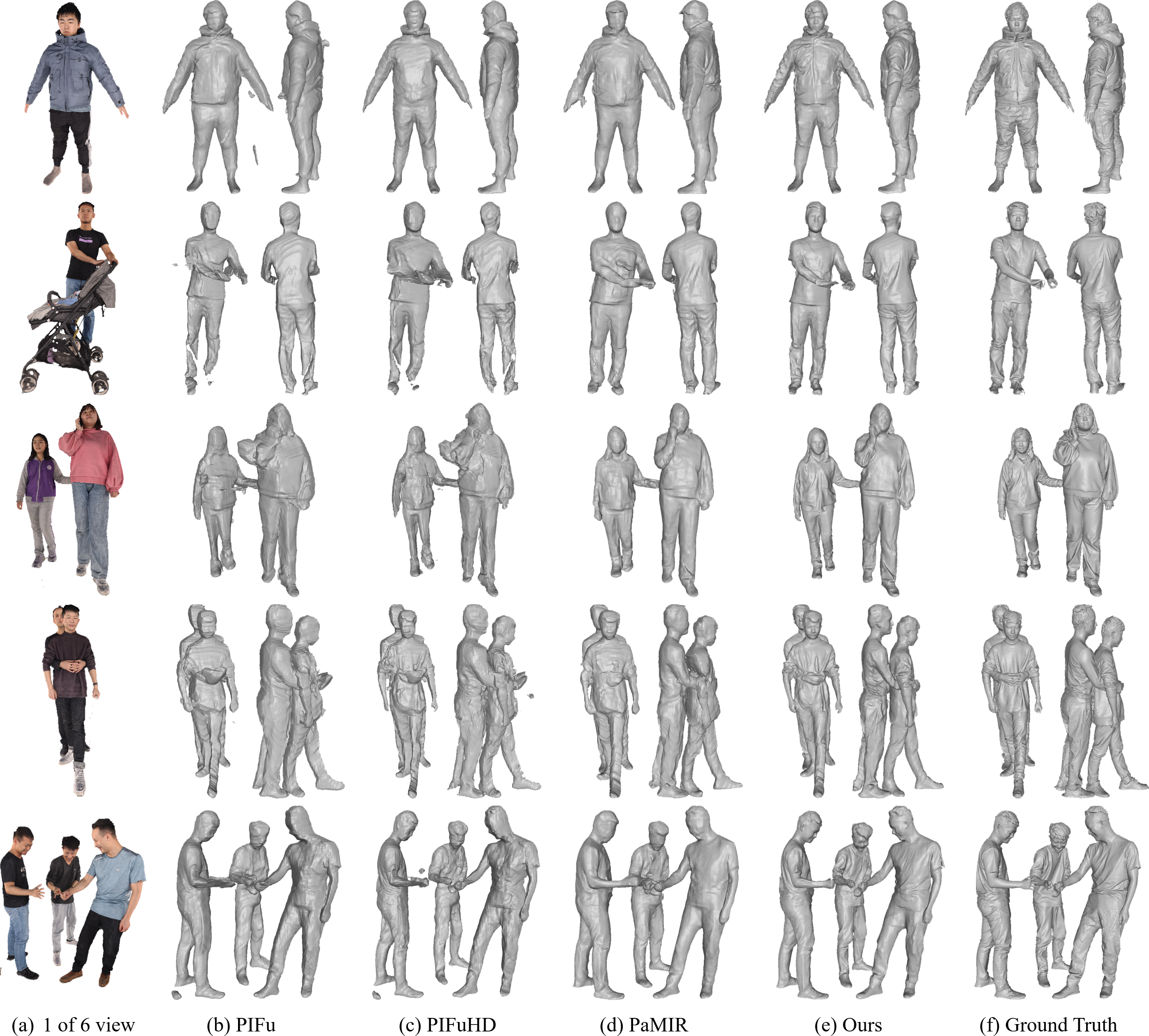}
   \vspace{-15pt}
   \caption{Reconstruction results on MultiHuman dataset of single person, occluded single person, 
   two natural-interactive person, two closely-interactive person, three person scenes (top to bottom). 
   Our method (e) generates robust and highly detailed humans, significantly narrowing the gap between ground truth (f) and performance of current 
   state-of-the-art methods. 
   }
   \label{fig:multi person}
   \vspace{-10pt}
\end{figure*}

\noindent \textbf{Performance on MultiHuman}
We compare our method with current state-of-the-art approaches, i.e, PIFu~\cite{saito2019pifu}, PIFuHD~\cite{saito2020pifuhd} and PaMIR~\cite{zheng2021pamir} (PIFu + SMPL).
All the methods are trained with the same setting as described in Sec.~\ref{sec 5}. 
For PIFuHD, the backside normal maps are not used in our implementation, and multi-view features are fused by mean pooling as~\cite{saito2019pifu, zheng2021pamir}. 
During test, the ground truth models are normalized to $180$ centimeters height and we render 6 view images as the input.
\begin{figure}[ht]
   \centering
   \vspace{-5pt}
   \includegraphics[width=\linewidth]{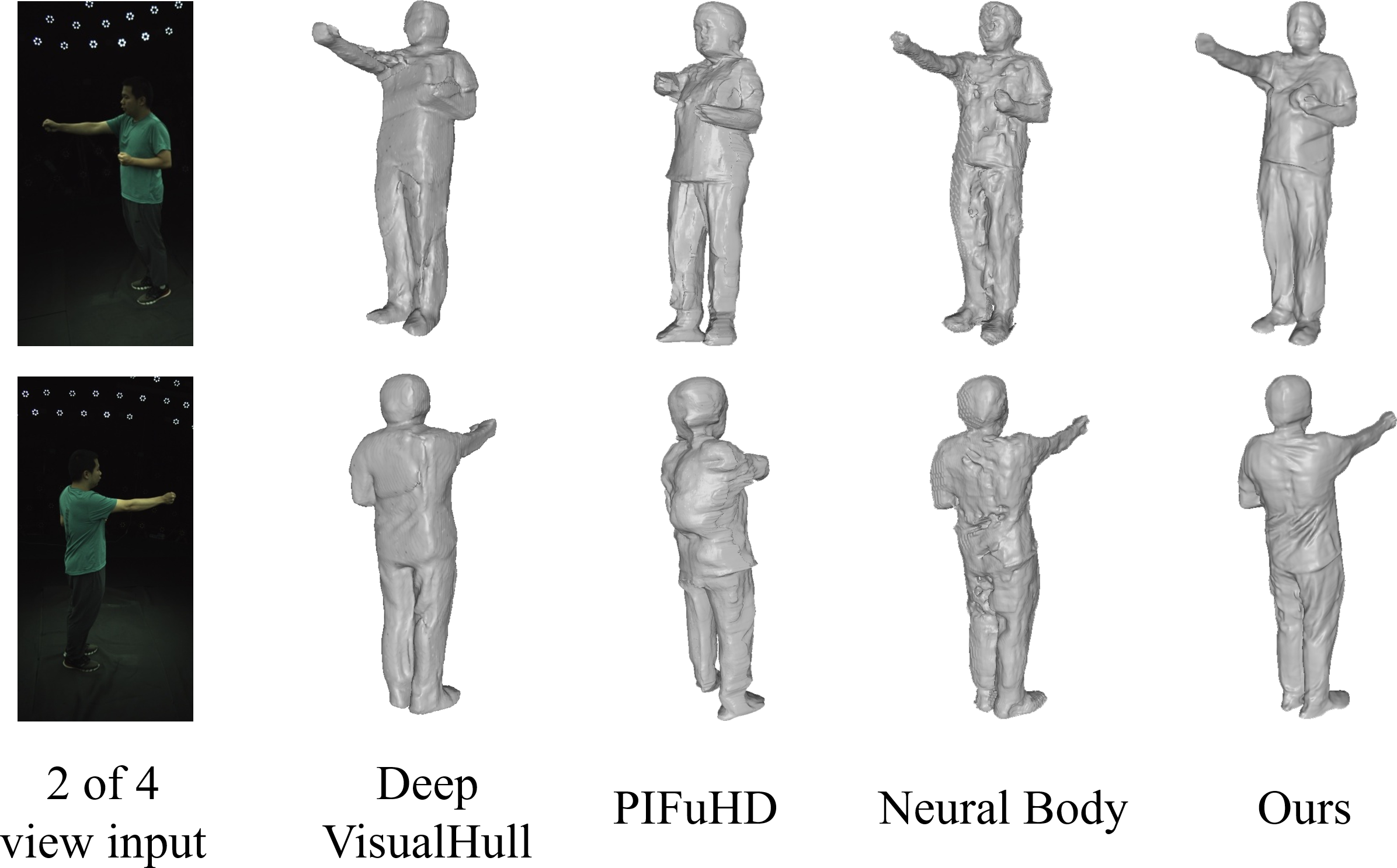}
   \vspace{-15pt}
   \caption{Performance on ZJU-Mocap dataset~\cite{peng2021neural}. Our method outperforms state-of-the-art approaches including DeepVisualHull~\cite{huang2018deep}, PIFuHD~\cite{saito2020pifuhd} and Neural Body~\cite{peng2021neural}.
   }
   \vspace{-10pt}
   \label{fig:nb}
\end{figure}
The point-to-surface distance and chamfer distance between the reconstruction and ground truth geometry are used as evaluation matrix. 
Quantitative results are shown in Table~\ref{tab:multiview}.
When occlusions intensify with increasing persons and interacting elements, the loss of prior methods exacerbate while ours remains competitive.
Qualitative results illustrated in Figure~\ref{fig:multi person} indicate the prominence of our method and the large gap between prior works and ours when handling occlusions in multi-person scenes. 
Our method is able to reconstruct highly detailed 3D human robustly even under closely interactive scenes.

\begin{table*}[ht!]
   \vspace{-5pt}
   \centering
   \begin{tabular}{lcccccccccc}
   \toprule
  \multirow{2}*{Method}  & \multicolumn{2}{c}{\makecell[c]{MultiHuman \\ \small{(single)}}} 
  & \multicolumn{2}{c}{\makecell[c]{MultiHuman \\ \small{(occluded single)}}} 
  & \multicolumn{2}{c}{\makecell[c]{MultiHuman \\ \small{(two natural-inter)}}} 
  & \multicolumn{2}{c}{\makecell[c]{MultiHuman \\ \small{(two closely-inter)}}} 
  & \multicolumn{2}{c}{\makecell[c]{MultiHuman \\ \small{(three)}}}\\
  & \footnotesize{Chamfer} & \footnotesize{P2S} & \footnotesize{Chamfer} & \footnotesize{P2S} & \footnotesize{Chamfer} & \footnotesize{P2S} & \footnotesize{Chamfer} & \footnotesize{P2S} & \footnotesize{Chamfer} & \footnotesize{P2S} \\ 
  \midrule
      PIFu \footnotesize{(Mview + Mean)}\cite{saito2019pifu}  & 1.131 & 1.220 & 1.402 & 1.522 & 1.578 & 1.620 &  1.745 & 1.831 & 1.780 & 1.564 \\ 
      \makecell[l]{PIFuHD \footnotesize{(Mview + Mean)}\cite{saito2020pifuhd}}  & 0.914 & 0.948 & 1.365 & 1.406 & 1.353 & 1.376 & 1.614 & 1.655 & 1.814 & 1.496 \\ 
      \makecell[l]{PAMIR \footnotesize{(Mview + Mean)}\cite{zheng2021pamir}} & 1.173 & 1.113 & 1.362 & 1.309 & 1.227 & 1.110 & 1.400 & 1.198 & 1.414 & 1.281 \\ 
   \midrule
      \makecell[l]{PIFu \footnotesize{(Mview + Att)}}  & 1.054 & 1.174 & 1.343 & 1.479 & 1.566 & 1.605 & 1.773 & 1.845 & 1.541 & 1.383 \\ 
      \makecell[l]{PIFuHD \footnotesize{(Mview + Att)}} & \textbf{0.845} & \textbf{0.867} &  1.195 & 1.189 & 1.278 & 1.272 & 1.515 & 1.450 & 1.468 & 1.287 \\
      \makecell[l]{Ours\footnotesize{(w/o Att)}}  & 1.015 & 0.967  & 1.251 & 1.181  & 1.020 & 0.925 & 1.264 & 1.088 & 1.309 & 1.246 \\
      \makecell[l]{Ours\footnotesize{(w/o SN)}}  & 1.063 & 1.017 & 1.277 & 1.233  & 1.126 & 0.989 & 1.357 & 1.141  & 1.334 & 1.155 \\ 
   \midrule
      Ours  & 0.895 & 0.887 & \textbf{1.041} & \textbf{1.021}  & \textbf{0.956} & \textbf{0.927} & \textbf{1.134} & \textbf{1.067} & \textbf{1.130} & \textbf{1.078} \\
   \bottomrule
\end{tabular}
   \caption{Quantitative evaluation on MultiHuman dataset. We compare our method with
   PIFu~\cite{saito2019pifu}, PIFuHD~\cite{saito2020pifuhd}, PaMIR~\cite{zheng2021pamir} with mean pooling feature fusion method~\cite{saito2019pifu}, and several variants including PIFu + Att (attention module), PIFuHD + Att, our method without attention (w/o att) and our method without the SMPL global normal maps (SN).}
  \vspace{-5pt}
   \label{tab:multiview}
  \end{table*}

\noindent \textbf{Performance on Real World Data}
We evaluate our method on ZJU-MoCap dataset~\cite{peng2021neural}, a multi-view real world dataset,
with comparison to DeepVisualHull~\cite{huang2018deep}, a volumetric performance capture from sparse multi-view, Neural Body~\cite{peng2021neural}, 
a differentiable rendering method directly trained on the test image sequence and PIFuHD~\cite{saito2020pifuhd}.
We re-implement DeepVisualHull and use the released code and pretrained models of PIFuHD and Neural Body .
Figure~\ref{fig:nb} shows the state-of-the-art performance of our method on the benchmark.
Reconstruction on real world images (6 view for our data and 8 view for TotalCapture dataset~\cite{joo2018total}) is demonstrated in Figure~\ref{teaser}. 

\begin{figure}[ht]
   \centering
   \vspace{15pt}
   \includegraphics[width=\linewidth]{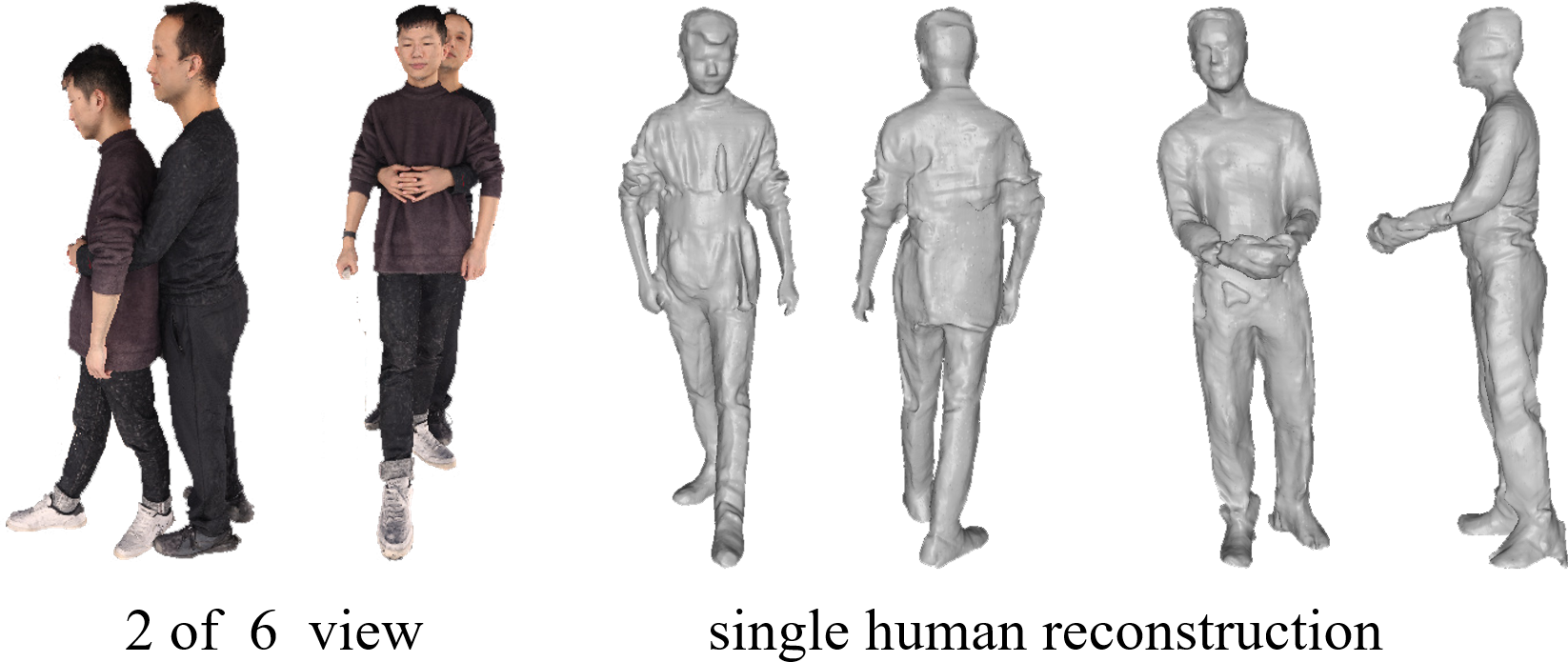}
   \vspace{-15pt}
   \caption{Our method generates robust results even when parts of human are not visible in closely interactive scenes.
   }
   \vspace{-25pt}
   \label{fig:occ}
\end{figure}

\begin{figure}[ht]
   \vspace{-5pt}
   \centering
   \includegraphics[width=\linewidth]{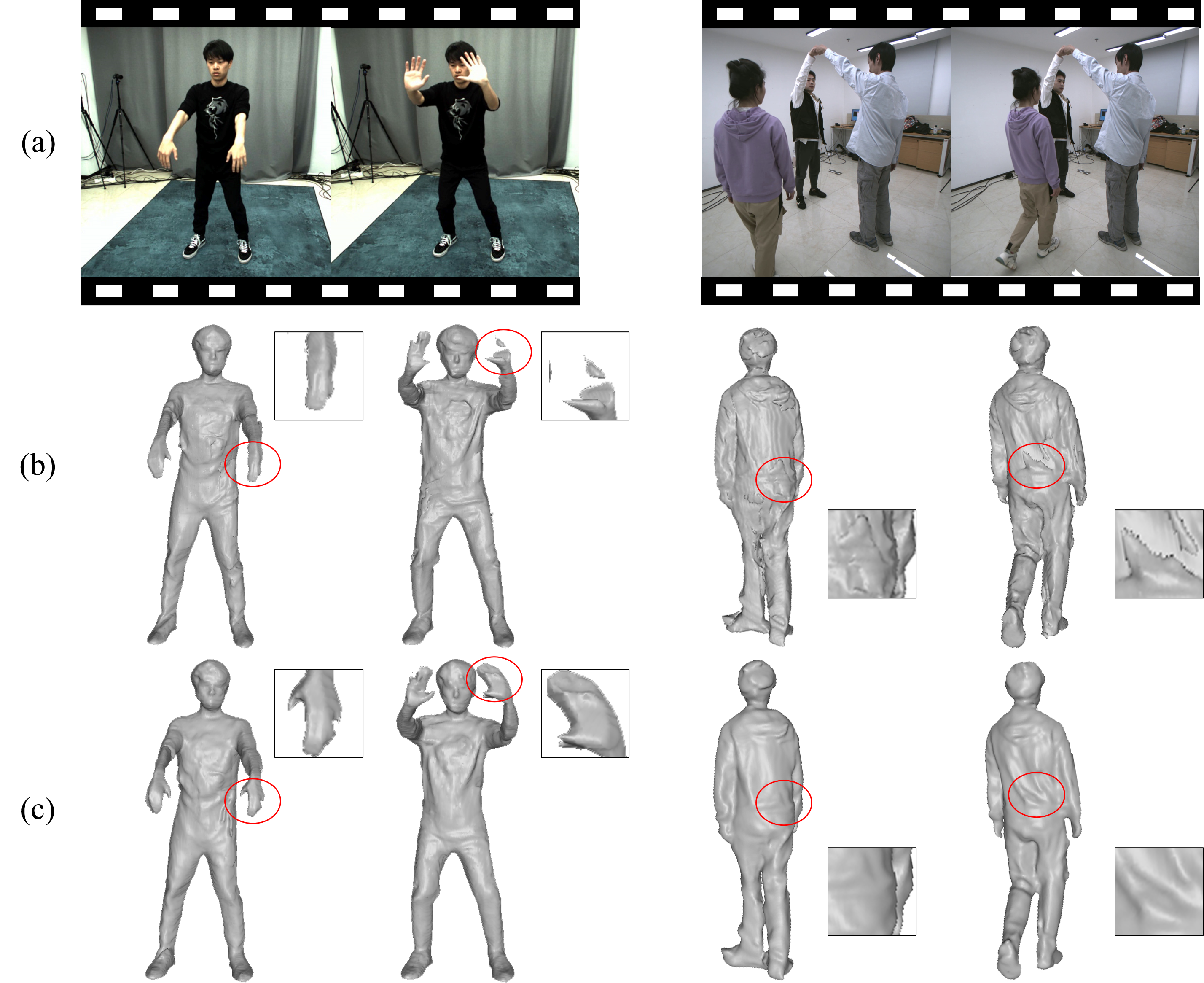}
   \vspace{-15pt}
   \caption{Performance on real world image sequence. (b) shows the original result of each frame, (c) demonstrates the results polished by temporal fusion.
   }
   \label{fig:video1}
   \vspace{-15pt}
\end{figure}

\subsection{Ablation Study}
This section aims to find the factors that contribute to the prominence of our method. 
We achieve the state-of-the-art performance mainly by 
leveraging a self-attention network combined with SMPL and a temporal fusion method for consistent results. 
We then demonstrate how the approaches improve the reconstruction under different situations.

\noindent \textbf{Variant 1: Self-attention Module} 
We design a self-attention module to better capture the details from different observations.
To figure out the strength of our multi-view feature fusion method, 
we combine the attention module with PIFu\cite{saito2019pifu} (PIFu + Att) and PIFuHD\cite{saito2020pifuhd} (PIFuHD + Att), 
and further evaluate our method's performance without the module (replaced by mean pooling). Quantitative results in Table~\ref{tab:multiview}
shows that the module benefits baseline models under non occluded and occluded scenes.
PIFuHD with attention module even outperforms ours on single human reconstruction, since the limitations brought by SMPL (Section~\ref{limitations}) can lead to a lower accuracy for our method.  
For PIFu the improvement is marginal, indicating that the module is more effective to 
merge multi-view features with the detailed geometry information offered by image normal maps.
For our method, we lose the competitive performance without the module.
Qualitative examples in Figure~\ref{fig:view consistency} further demonstrate how the module can help the baseline model maintain geometry details with increasing views.

\noindent \textbf{Variant 2: Use of SMPL}
SMPL is used in our method as a 3D proxy for the network to generate a reasonable output, 
and we further design a SMPL global normal map (described in Section~\ref{sec:smpl}) to improve the robustness of reconstruction against occlusions and preserving details.
The huge gap between PaMIR~\cite{zheng2021pamir} and ours indicates SMPL is not only the factor contributing to our advantages. 
Table~\ref{tab:multiview} shows the performance of our method without the designed global maps (Ours w/o SN). 
The results demonstrate lower accuracy of reconstruction, 
which implies the efficiency of the global maps as a visual reference to guide the attention network merge multi-view information.

\noindent \textbf{Variant 3: Temporal Fusion} 
Figure~\ref{fig:video1} illustrates the results of our method with and without temporal fusion on real world image sequence. 
The temporal fusion method further enhance the reconstruction consistency, which can be witnessed more clearly in our supplementary video.

\noindent \textbf{Robustness against occlusions}
Quantitative results (Table~\ref{tab:multiview}) shows that our method maintains high accuracy with increasing occlusions.
Figure~\ref{fig:occ} illustrates that when the body parts are invisible due to heavy occlusion, smooth results will be generated.
 
\subsection{Limitations}
\label{limitations}
\begin{figure}[ht]
   \centering
   \vspace{-5pt}
   \includegraphics[width=\linewidth]{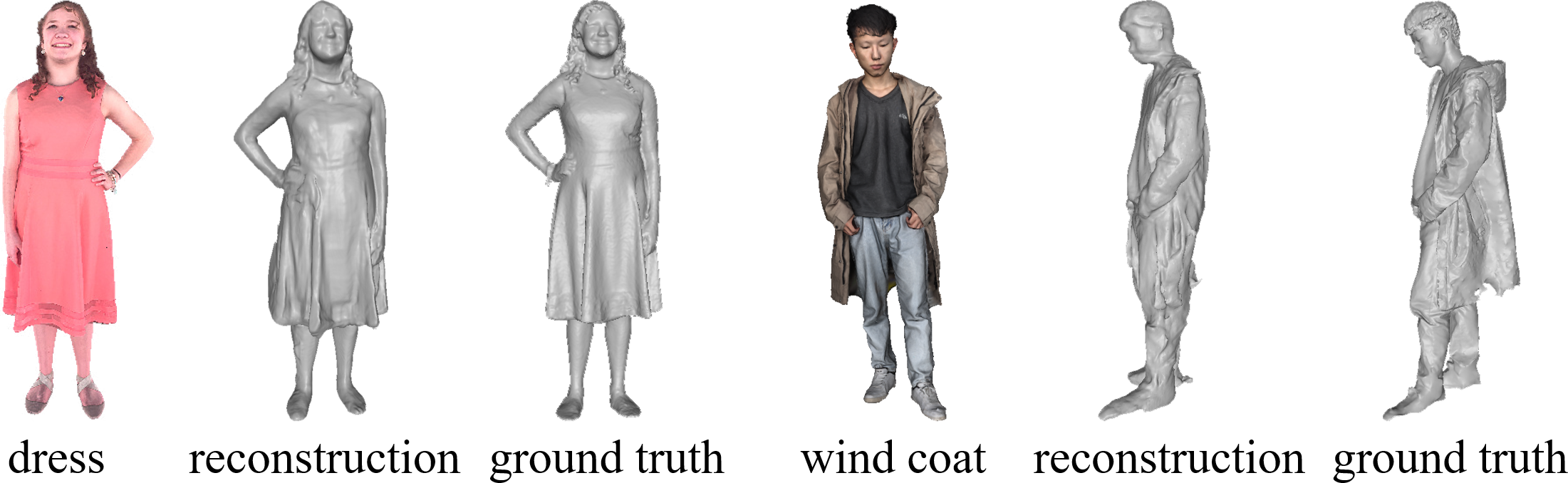}
   \vspace{-15pt}
   \caption{Reconstruction of challenging clothes.
   }
   \label{fig:clothes}
   \vspace{-10pt}
\end{figure}

Since we use SMPL as a 3D reference, our method cannot reconstruction other objects aside from human. 
For challenging clothes, Figure~\ref{fig:clothes} demonstrates that we are able to reconstruct tight dress, while for loose clothing like a wind coat, the reconstruction can be unstable.

\begin{figure}[ht]
   \centering   
   \includegraphics[width=\linewidth]{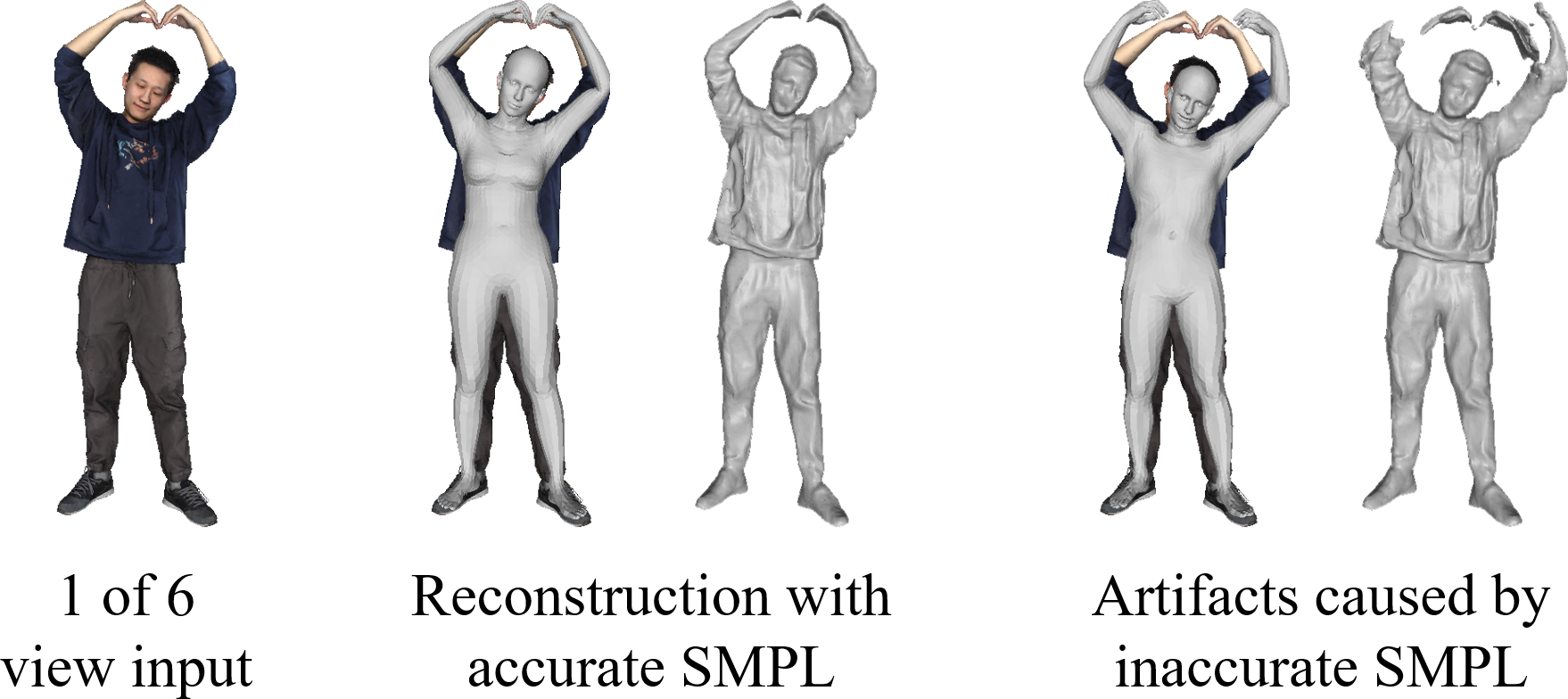}
   \vspace{-15pt}
   \caption{Failure case with inaccurate SMPL input. Our method is misled by the incorrect SMPL information.}
   \vspace{-15pt}
   \label{fig:inaccurate smpl}
\end{figure}

Besides, our method relies on a well fitted SMPL, i.e, the SMPL body within the correct region. An inaccurate SMPL can lead to artifacts and failure cases (Figure~\ref{fig:inaccurate smpl}).

\section{Discussion and Future Works}
Our method is capable of reconstructing high-fidelity multi-person with the suppose that the preprocessing is well conditioned, i.e., calibrated cameras, accurate segmentation, well fitted SMPL.
Though in real world cases the system remains robust against small noise, large errors in preprocessing could lead to failure cases.
Future works can focus on a more simplified pipeline, e.g., autonomous calibration, reconstruction using implicit human template, 
and lightweight networks to achieve real-time inference, which will surely make the system more applicable.
 
\noindent\textbf{Acknowledgements} This work is supported by the National Key Research and Development Program of China No.2018YFB2100500, the NSFC No.61827805, the NSFC No.62125107, the NSFC No.62171255 and Shuimu Tsinghua Scholarship.

{\small
\bibliographystyle{ieee_fullname}
\bibliography{egbib}

\begin{thebibliography}{10}\itemsep=-1pt

\bibitem{alldieck2019tex2shape}
Thiemo Alldieck, Gerard Pons-Moll, Christian Theobalt, and Marcus Magnor.
\newblock Tex2shape: Detailed full human body geometry from a single image.
\newblock In {\em ICCV}, 2019.

\bibitem{bagautdinov2021driving}
Timur Bagautdinov, Chenglei Wu, Tomas Simon, Fabian Prada, Takaaki Shiratori,
  Shih-En Wei, Weipeng Xu, Yaser Sheikh, and Jason Saragih.
\newblock Driving-signal aware full-body avatars.
\newblock {\em ACM Transactions on Graphics (TOG)}, 40(4):1--17, 2021.

\bibitem{belagiannis20143d}
Vasileios Belagiannis, Sikandar Amin, Mykhaylo Andriluka, Bernt Schiele, Nassir
  Navab, and Slobodan Ilic.
\newblock 3d pictorial structures for multiple human pose estimation.
\newblock In {\em CVPR}, pages 1669--1676, 2014.

\bibitem{bhatnagar2020combining}
Bharat~Lal Bhatnagar, Cristian Sminchisescu, Christian Theobalt, and Gerard
  Pons-Moll.
\newblock Combining implicit function learning and parametric models for 3d
  human reconstruction.
\newblock In {\em ECCV}, pages 311--329. Springer, 2020.

\bibitem{bhatnagar2020loopreg}
Bharat~Lal Bhatnagar, Cristian Sminchisescu, Christian Theobalt, and Gerard
  Pons-Moll.
\newblock Loopreg: Self-supervised learning of implicit surface
  correspondences, pose and shape for 3d human mesh registration.
\newblock {\em NeurIPS}, 33, 2020.

\bibitem{bridgeman2019multi}
Lewis Bridgeman, Marco Volino, Jean-Yves Guillemaut, and Adrian Hilton.
\newblock Multi-person 3d pose estimation and tracking in sports.
\newblock In {\em CVPR}, 2019.

\bibitem{chibane2020implicit}
Julian Chibane, Thiemo Alldieck, and Gerard Pons-Moll.
\newblock Implicit functions in feature space for 3d shape reconstruction and
  completion.
\newblock In {\em CVPR}, pages 6970--6981, 2020.

\bibitem{collet2015high}
Alvaro Collet, Ming Chuang, Pat Sweeney, Don Gillett, Dennis Evseev, David
  Calabrese, Hugues Hoppe, Adam Kirk, and Steve Sullivan.
\newblock High-quality streamable free-viewpoint video.
\newblock {\em ACM Transactions on Graphics (ToG)}, 34(4):1--13, 2015.

\bibitem{dai2019second}
Tao Dai, Jianrui Cai, Yongbing Zhang, Shu-Tao Xia, and Lei Zhang.
\newblock Second-order attention network for single image super-resolution.
\newblock In {\em CVPR}, pages 11065--11074, 2019.

\bibitem{de2008performance}
Edilson De~Aguiar, Carsten Stoll, Christian Theobalt, Naveed Ahmed, Hans-Peter
  Seidel, and Sebastian Thrun.
\newblock Performance capture from sparse multi-view video.
\newblock In {\em ACM SIGGRAPH 2008 papers}, pages 1--10. 2008.

\bibitem{deng2020nasa}
Boyang Deng, John~P Lewis, Timothy Jeruzalski, Gerard Pons-Moll, Geoffrey
  Hinton, Mohammad Norouzi, and Andrea Tagliasacchi.
\newblock Nasa neural articulated shape approximation.
\newblock In {\em ECCV}, pages 612--628. Springer, 2020.

\bibitem{dong2019fast}
Junting Dong, Wen Jiang, Qixing Huang, Hujun Bao, and Xiaowei Zhou.
\newblock Fast and robust multi-person 3d pose estimation from multiple views.
\newblock In {\em CVPR}, pages 7792--7801, 2019.

\bibitem{dou2017motion2fusion}
Mingsong Dou, Philip Davidson, Sean~Ryan Fanello, Sameh Khamis, Adarsh Kowdle,
  Christoph Rhemann, Vladimir Tankovich, and Shahram Izadi.
\newblock Motion2fusion: Real-time volumetric performance capture.
\newblock {\em ACM Transactions on Graphics (TOG)}, 36(6):1--16, 2017.

\bibitem{dou2013scanning}
Mingsong Dou, Henry Fuchs, and Jan-Michael Frahm.
\newblock Scanning and tracking dynamic objects with commodity depth cameras.
\newblock In {\em 2013 IEEE international symposium on mixed and augmented
  Reality (ISMAR)}, pages 99--106. IEEE, 2013.

\bibitem{dou2016fusion4d}
Mingsong Dou, Sameh Khamis, Yury Degtyarev, Philip Davidson, Sean~Ryan Fanello,
  Adarsh Kowdle, Sergio~Orts Escolano, Christoph Rhemann, David Kim, Jonathan
  Taylor, et~al.
\newblock Fusion4d: Real-time performance capture of challenging scenes.
\newblock {\em ACM Transactions on Graphics (TOG)}, 35(4):1--13, 2016.

\bibitem{naseer2021intriguing}
Naseer~Muzammal et al.
\newblock Intriguing properties of vision transformers.
\newblock {\em arXiv preprint arXiv:2105.10497}, 2021.

\bibitem{gabeur2019moulding}
Valentin Gabeur, Jean-S{\'e}bastien Franco, Xavier Martin, Cordelia Schmid, and
  Gregory Rogez.
\newblock Moulding humans: Non-parametric 3d human shape estimation from single
  images.
\newblock In {\em ICCV}, pages 2232--2241, 2019.

\bibitem{gall2009motion}
Juergen Gall, Carsten Stoll, Edilson De~Aguiar, Christian Theobalt, Bodo
  Rosenhahn, and Hans-Peter Seidel.
\newblock Motion capture using joint skeleton tracking and surface estimation.
\newblock In {\em CVPR}, pages 1746--1753. IEEE, 2009.

\bibitem{Minimal18}
Andrew Gilbert, Marco Volino, John~P. Collomosse, and Adrian Hilton.
\newblock Volumetric performance capture from minimal camera viewpoints.
\newblock In {\em ECCV}, volume 11215, pages 591--607. Springer, 2018.

\bibitem{guo2019relightables}
Kaiwen Guo, Peter Lincoln, Philip Davidson, Jay Busch, Xueming Yu, Matt Whalen,
  Geoff Harvey, Sergio Orts-Escolano, Rohit Pandey, Jason Dourgarian, et~al.
\newblock The relightables: Volumetric performance capture of humans with
  realistic relighting.
\newblock {\em ACM Transactions on Graphics (TOG)}, 38(6):1--19, 2019.

\bibitem{habermann2020deepcap}
Marc Habermann, Weipeng Xu, Michael Zollhofer, Gerard Pons-Moll, and Christian
  Theobalt.
\newblock Deepcap: Monocular human performance capture using weak supervision.
\newblock In {\em CVPR}, pages 5052--5063, 2020.

\bibitem{he2021challencap}
Yannan He, Anqi Pang, Xin Chen, Han Liang, Minye Wu, Yuexin Ma, and Lan Xu.
\newblock Challencap: Monocular 3d capture of challenging human performances
  using multi-modal references.
\newblock In {\em CVPR}, pages 11400--11411, 2021.

\bibitem{hu2019taichi}
Yuanming Hu, Tzu-Mao Li, Luke Anderson, Jonathan Ragan-Kelley, and Fr{\'e}do
  Durand.
\newblock Taichi: a language for high-performance computation on spatially
  sparse data structures.
\newblock {\em ACM Transactions on Graphics (TOG)}, 38(6):201, 2019.

\bibitem{huang2020hot}
Lin Huang, Jianchao Tan, Jingjing Meng, Ji Liu, and Junsong Yuan.
\newblock Hot-net: Non-autoregressive transformer for 3d hand-object pose
  estimation.
\newblock In {\em Proceedings of the 28th ACM International Conference on
  Multimedia}, pages 3136--3145, 2020.

\bibitem{huang2018deep}
Zeng Huang, Tianye Li, Weikai Chen, Yajie Zhao, Jun Xing, Chloe LeGendre,
  Linjie Luo, Chongyang Ma, and Hao Li.
\newblock Deep volumetric video from very sparse multi-view performance
  capture.
\newblock In {\em ECCV}, pages 336--354, 2018.

\bibitem{huang2020arch}
Zeng Huang, Yuanlu Xu, Christoph Lassner, Hao Li, and Tony Tung.
\newblock Arch: Animatable reconstruction of clothed humans.
\newblock In {\em CVPR}, pages 3093--3102, 2020.

\bibitem{Joo_2015_ICCV}
Hanbyul Joo, Hao Liu, Lei Tan, Lin Gui, Bart Nabbe, Iain Matthews, Takeo
  Kanade, Shohei Nobuhara, and Yaser Sheikh.
\newblock Panoptic studio: A massively multiview system for social motion
  capture.
\newblock In {\em ICCV}, 2015.

\bibitem{joo2019panoptic}
H Joo, T Simon, X Li, H Liu, L Tan, L Gui, S Banerjee, T Godisart, B Nabbe, I
  Matthews, et~al.
\newblock Panoptic studio: A massively multiview system for social interaction
  capture.
\newblock {\em TPAMI}, 41(1):190--204, 2019.

\bibitem{joo2018total}
Hanbyul Joo, Tomas Simon, and Yaser Sheikh.
\newblock Total capture: A 3d deformation model for tracking faces, hands, and
  bodies.
\newblock In {\em CVPR}, pages 8320--8329. IEEE, 2018.

\bibitem{kwon2020recursive}
Oh-Hun Kwon, Julian Tanke, and Juergen Gall.
\newblock Recursive bayesian filtering for multiple human pose tracking from
  multiple cameras.
\newblock In {\em Proceedings of the Asian Conference on Computer Vision},
  2020.

\bibitem{li2020self}
Peike Li, Yunqiu Xu, Yunchao Wei, and Yi Yang.
\newblock Self-correction for human parsing.
\newblock {\em IEEE TPAMI}, 2020.

\bibitem{Monoport2020}
Ruilong Li, Yuliang Xiu, Shunsuke Saito, Zeng Huang, Kyle Olszewski, and Hao
  Li.
\newblock Monocular real-time volumetric performance capture.
\newblock In {\em ECCV}, 2020.

\bibitem{li2019attention}
Yanwei Li, Xinze Chen, Zheng Zhu, Lingxi Xie, Guan Huang, Dalong Du, and
  Xingang Wang.
\newblock Attention-guided unified network for panoptic segmentation.
\newblock In {\em CVPR}, pages 7026--7035, 2019.

\bibitem{lin2021multi}
Jiahao Lin and Gim~Hee Lee.
\newblock Multi-view multi-person 3d pose estimation with plane sweep stereo.
\newblock In {\em CVPR}, pages 11886--11895, 2021.

\bibitem{liu2013markerless}
Yebin Liu, Juergen Gall, Carsten Stoll, Qionghai Dai, Hans-Peter Seidel, and
  Christian Theobalt.
\newblock Markerless motion capture of multiple characters using multiview
  image segmentation.
\newblock {\em TPAMI}, 35(11):2720--2735, 2013.

\bibitem{liu2011markerless}
Yebin Liu, Carsten Stoll, Juergen Gall, Hans-Peter Seidel, and Christian
  Theobalt.
\newblock Markerless motion capture of interacting characters using multi-view
  image segmentation.
\newblock In {\em CVPR}, pages 1249--1256. IEEE, 2011.

\bibitem{loper2015smpl}
Matthew Loper, Naureen Mahmood, Javier Romero, Gerard Pons-Moll, and Michael~J
  Black.
\newblock Smpl: A skinned multi-person linear model.
\newblock {\em ACM transactions on graphics (TOG)}, 34(6):1--16, 2015.

\bibitem{luo2020attention}
Keyang Luo, Tao Guan, Lili Ju, Yuesong Wang, Zhuo Chen, and Yawei Luo.
\newblock Attention-aware multi-view stereo.
\newblock In {\em CVPR}, pages 1590--1599, 2020.

\bibitem{mehta2020xnect}
Dushyant Mehta, Oleksandr Sotnychenko, Franziska Mueller, Weipeng Xu, Mohamed
  Elgharib, Pascal Fua, Hans-Peter Seidel, Helge Rhodin, Gerard Pons-Moll, and
  Christian Theobalt.
\newblock Xnect: Real-time multi-person 3d motion capture with a single rgb
  camera.
\newblock {\em ACM Transactions on Graphics (TOG)}, 39(4):82--1, 2020.

\bibitem{natsume2019siclope}
Ryota Natsume, Shunsuke Saito, Zeng Huang, Weikai Chen, Chongyang Ma, Hao Li,
  and Shigeo Morishima.
\newblock Siclope: Silhouette-based clothed people.
\newblock In {\em CVPR}, pages 4480--4490, 2019.

\bibitem{ohashi2020synergetic}
Takuya Ohashi, Yosuke Ikegami, and Yoshihiko Nakamura.
\newblock Synergetic reconstruction from 2d pose and 3d motion for wide-space
  multi-person video motion capture in the wild.
\newblock {\em Image and Vision Computing}, 104:104028, 2020.

\bibitem{pang2021few}
Anqi Pang, Xin Chen, Haimin Luo, Minye Wu, Jingyi Yu, and Lan Xu.
\newblock Few-shot neural human performance rendering from sparse rgbd videos.
\newblock {\em IJCAI}, 2021.

\bibitem{pavlakos2019expressive}
Georgios Pavlakos, Vasileios Choutas, Nima Ghorbani, Timo Bolkart, Ahmed~AA
  Osman, Dimitrios Tzionas, and Michael~J Black.
\newblock Expressive body capture: 3d hands, face, and body from a single
  image.
\newblock In {\em CVPR}, pages 10975--10985, 2019.

\bibitem{peng2021neural}
Sida Peng, Yuanqing Zhang, Yinghao Xu, Qianqian Wang, Qing Shuai, Hujun Bao,
  and Xiaowei Zhou.
\newblock Neural body: Implicit neural representations with structured latent
  codes for novel view synthesis of dynamic humans.
\newblock In {\em CVPR}, 2021.

\bibitem{rogez2017lcr}
Gregory Rogez, Philippe Weinzaepfel, and Cordelia Schmid.
\newblock Lcr-net: Localization-classification-regression for human pose.
\newblock In {\em CVPR}, pages 3433--3441, 2017.

\bibitem{rogez2019lcr}
Gregory Rogez, Philippe Weinzaepfel, and Cordelia Schmid.
\newblock Lcr-net++: Multi-person 2d and 3d pose detection in natural images.
\newblock {\em IEEE TPAMI}, 42(5):1146--1161, 2019.

\bibitem{saito2019pifu}
Shunsuke Saito, Zeng Huang, Ryota Natsume, Shigeo Morishima, Angjoo Kanazawa,
  and Hao Li.
\newblock Pifu: Pixel-aligned implicit function for high-resolution clothed
  human digitization.
\newblock In {\em ICCV}, October 2019.

\bibitem{saito2020pifuhd}
Shunsuke Saito, Tomas Simon, Jason Saragih, and Hanbyul Joo.
\newblock Pifuhd: Multi-level pixel-aligned implicit function for
  high-resolution 3d human digitization.
\newblock In {\em CVPR}, 2020.

\bibitem{shao2021doublefield}
Ruizhi Shao, Hongwen Zhang, He Zhang, Yanpei Cao, Tao Yu, and Yebin Liu.
\newblock Doublefield: Bridging the neural surface and radiance fields for
  high-fidelity human rendering.
\newblock {\em arXiv preprint arXiv:2106.03798}, 2021.

\bibitem{smith2019facsimile}
David Smith, Matthew Loper, Xiaochen Hu, Paris Mavroidis, and Javier Romero.
\newblock Facsimile: Fast and accurate scans from an image in less than a
  second.
\newblock In {\em ICCV}, pages 5330--5339, 2019.

\bibitem{Starck07}
Jonathan Starck and Adrian Hilton.
\newblock Surface capture for performance-based animation.
\newblock {\em {IEEE} Computer Graphics and Applications}, 27(3):21--31, 2007.

\bibitem{tu2020voxelpose}
Hanyue Tu, Chunyu Wang, and Wenjun Zeng.
\newblock Voxelpose: Towards multi-camera 3d human pose estimation in wild
  environment.
\newblock In {\em ECCV}, pages 197--212. Springer, 2020.

\bibitem{varol2018bodynet}
Gul Varol, Duygu Ceylan, Bryan Russell, Jimei Yang, Ersin Yumer, Ivan Laptev,
  and Cordelia Schmid.
\newblock Bodynet: Volumetric inference of 3d human body shapes.
\newblock In {\em ECCV}, pages 20--36, 2018.

\bibitem{vaswani2017attention}
Ashish Vaswani, Noam Shazeer, Niki Parmar, Jakob Uszkoreit, Llion Jones,
  Aidan~N Gomez, {\L}ukasz Kaiser, and Illia Polosukhin.
\newblock Attention is all you need.
\newblock In {\em NeurIPS}, pages 5998--6008, 2017.

\bibitem{vlasic2008articulated}
Daniel Vlasic, Ilya Baran, Wojciech Matusik, and Jovan Popovi{\'c}.
\newblock Articulated mesh animation from multi-view silhouettes.
\newblock In {\em ACM SIGGRAPH 2008 papers}, pages 1--9. 2008.

\bibitem{vlasic2009dynamic}
Daniel Vlasic, Pieter Peers, Ilya Baran, Paul Debevec, Jovan Popovi{\'c},
  Szymon Rusinkiewicz, and Wojciech Matusik.
\newblock Dynamic shape capture using multi-view photometric stereo.
\newblock In {\em ACM SIGGRAPH Asia 2009 papers}, pages 1--11. 2009.

\bibitem{wang2017residual}
Fei Wang, Mengqing Jiang, Chen Qian, Shuo Yang, Cheng Li, Honggang Zhang,
  Xiaogang Wang, and Xiaoou Tang.
\newblock Residual attention network for image classification.
\newblock In {\em CVPR}, pages 3156--3164, 2017.

\bibitem{Wang_2021_CVPR}
Ziyan Wang, Timur Bagautdinov, Stephen Lombardi, Tomas Simon, Jason Saragih,
  Jessica Hodgins, and Michael Zollhofer.
\newblock Learning compositional radiance fields of dynamic human heads.
\newblock In {\em CVPR}, pages 5704--5713, June 2021.

\bibitem{yu2018learning}
Changqian Yu, Jingbo Wang, Chao Peng, Changxin Gao, Gang Yu, and Nong Sang.
\newblock Learning a discriminative feature network for semantic segmentation.
\newblock In {\em CVPR}, pages 1857--1866, 2018.

\bibitem{yu2017bodyfusion}
Tao Yu, Kaiwen Guo, Feng Xu, Yuan Dong, Zhaoqi Su, Jianhui Zhao, Jianguo Li,
  Qionghai Dai, and Yebin Liu.
\newblock Bodyfusion: Real-time capture of human motion and surface geometry
  using a single depth camera.
\newblock In {\em ICCV}, pages 910--919, 2017.

\bibitem{yu2021function4d}
Tao Yu, Zerong Zheng, Kaiwen Guo, Pengpeng Liu, Qionghai Dai, and Yebin Liu.
\newblock Function4d: Real-time human volumetric capture from very sparse
  consumer rgbd sensors.
\newblock In {\em CVPR}, pages 5746--5756, 2021.

\bibitem{yu2018doublefusion}
Tao Yu, Zerong Zheng, Kaiwen Guo, Jianhui Zhao, Qionghai Dai, Hao Li, Gerard
  Pons-Moll, and Yebin Liu.
\newblock Doublefusion: Real-time capture of human performances with inner body
  shapes from a single depth sensor.
\newblock In {\em ICCV}, pages 7287--7296, 2018.

\bibitem{Zanfir_2018_CVPR}
Andrei Zanfir, Elisabeta Marinoiu, and Cristian Sminchisescu.
\newblock Monocular 3d pose and shape estimation of multiple people in natural
  scenes - the importance of multiple scene constraints.
\newblock In {\em CVPR}, June 2018.

\bibitem{zhang2018context}
Hang Zhang, Kristin Dana, Jianping Shi, Zhongyue Zhang, Xiaogang Wang, Ambrish
  Tyagi, and Amit Agrawal.
\newblock Context encoding for semantic segmentation.
\newblock In {\em CVPR}, pages 7151--7160, 2018.

\bibitem{pymaf2021}
Hongwen Zhang, Yating Tian, Xinchi Zhou, Wanli Ouyang, Yebin Liu, Limin Wang,
  and Zhenan Sun.
\newblock Pymaf: 3d human pose and shape regression with pyramidal mesh
  alignment feedback loop.
\newblock In {\em ICCV}, 2021.

\bibitem{zhang20204d}
Yuxiang Zhang, Liang An, Tao Yu, Xiu Li, Kun Li, and Yebin Liu.
\newblock 4d association graph for realtime multi-person motion capture using
  multiple video cameras.
\newblock In {\em CVPR}, pages 1324--1333, 2020.

\bibitem{lightcap2021}
Yuxiang Zhang, Zhe Li, Tao Yu, Liang Li, Mengcheng and~An, and Yebin Liu.
\newblock Light-weight multi-person total capture using sparse multi-view
  cameras.
\newblock In {\em ICCV}, 2021.

\bibitem{zheng2021pamir}
Zerong Zheng, Tao Yu, Yebin Liu, and Qionghai Dai.
\newblock Pamir: Parametric model-conditioned implicit representation for
  image-based human reconstruction.
\newblock {\em IEEE TPAMI}, 2021.

\bibitem{zheng2019deephuman}
Zerong Zheng, Tao Yu, Yixuan Wei, Qionghai Dai, and Yebin Liu.
\newblock Deephuman: 3d human reconstruction from a single image.
\newblock In {\em ICCV}, pages 7739--7749, 2019.

\bibitem{zhi2020texmesh}
Tiancheng Zhi, Christoph Lassner, Tony Tung, Carsten Stoll, Srinivasa~G
  Narasimhan, and Minh Vo.
\newblock Texmesh: Reconstructing detailed human texture and geometry from
  rgb-d video.
\newblock In {\em ECCV}, pages 492--509. Springer, 2020.

\bibitem{zhu2019detailed}
Hao Zhu, Xinxin Zuo, Sen Wang, Xun Cao, and Ruigang Yang.
\newblock Detailed human shape estimation from a single image by hierarchical
  mesh deformation.
\newblock In {\em CVPR}, pages 4491--4500, 2019.

\end{thebibliography}
}
\end{document}